## Viewpoint

# Harnessing ChatGPT for thematic analysis: Are we ready?

V Vien Lee[1,#], PhD; Stephanie C. C. van der Lubbe[1,#], PhD; Lay Hoon Goh[1, 2], MD; Jose M. Valderas[1,2,3], Prof, MD, PhD

[1] Division of Family Medicine, Yong Loo Lin School of Medicine, National University of Singapore, Singapore
[2] Department of Family Medicine, National University Health System, Singapore, Singapore
[3] Centre for Health Services Research, National University Health System, Singapore, Singapore
[#] Authors contributed equally to this work.

**Corresponding author:**
Jose M. Valderas, Prof, MD, PhD
Phone: +65 6772 3874
Email: jmvalderas@nus.edu.sg

## Abstract

ChatGPT is an advanced natural language processing tool with growing applications across various disciplines in medical research. Thematic analysis, a qualitative research method to identify and interpret patterns in data, is one application that stands to benefit from this technology. This viewpoint explores the utilization of ChatGPT in three core phases of thematic analysis within a medical context: 1) direct coding of transcripts, 2) generating themes from a predefined list of codes, and 3) preprocessing quotes for manuscript inclusion. Additionally, we explore the potential of ChatGPT to generate interview transcripts, which may be used for training purposes. We assess the strengths and limitations of using ChatGPT in these roles, highlighting areas where human intervention remains necessary. Overall, we argue that ChatGPT can function as a valuable tool during analysis, enhancing the efficiency of the thematic analysis and offering additional insights into the qualitative data.

## Introduction

Thematic analysis is a method to analyze qualitative data, commonly obtained through semi-structured interviews or focus groups, with the aim of identifying and interpreting patterns of meaning or themes within the data [1]. As a method, thematic analysis is inherently flexible and dependent on the researcher's underlying philosophical assumptions [2]. For instance, positivist approaches may place greater emphasis on coding reliability, while interpretivist approaches may place more significance on reflexivity and the researcher's role (including subjectivity) in knowledge production [2]. Accordingly, thematic analysis may be well-suited to meet varying research needs and requirements [3]. While there are multiple methods for thematic analysis, Braun and Clarke's six phases of thematic analysis is one of the most widely used approaches (see Figure 1) [1].

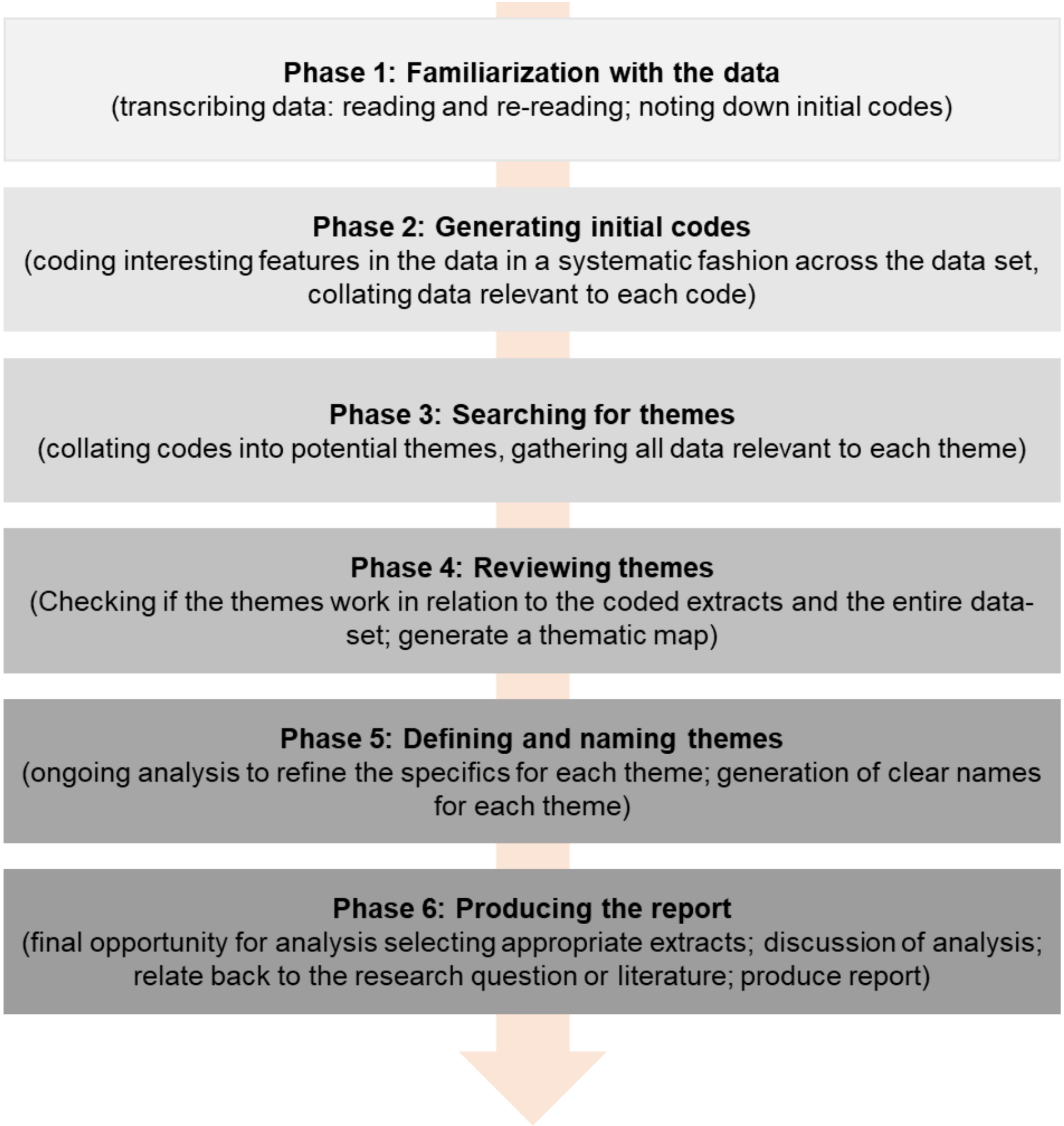

**Figure 1.** Braun and Clarke's six-phases of thematic analysis [1].

Given the flexibility of thematic analysis, there is room for creativity when engaging with the data and exploring tools that may aid the researcher's analytic process. With the increasing adoption of natural language processing (NLP) in healthcare research, such as diagnostic evaluation of electronic health records and the prediction of clinical outcomes based on consultation notes [4-6], researchers have begun to explore if there is space for Artificial

Intelligence (AI) within the domain of qualitative research. To date, several AI-based tools, such as AILYZE and MonkeyLearn, are available to aid researchers in conducting thematic analysis [7, 8]. For instance, AILYZE is able to summarize interview transcripts, provide suggestions for themes, and extract relevant quotes for each theme [7]. Nevertheless, full access to these tools often requires subscription payments.

In November 2022, OpenAI released version 3.5 of Chat Generative Pre-trained Transformer (ChatGPT-3.5), a large language model-based chatbot capable of performing a wide range of text-based tasks based on context and past conversations (e.g., summarizing research articles, answering domain-specific questions, generating outlines for manuscripts) [9]. ChatGPT-3.5 is the chatbot adaptation of GPT-3.5 and is specifically optimized for interactive conversations, though GPT-3.5 and ChatGPT-3.5 share the same foundational model. ChatGPT-3.5 is able to process a request and provide a response within a combined limit of 4,096 tokens (i.e. textual units, equivalent to approximately 3,000 words in English), typically within a few seconds and free of charge [9, 10].

Due to its availability to the public and free-to-use model, there has been a proliferation of discussion in the scientific community about incorporating GPT and ChatGPT into various aspects of research, including literature review, data processing, and manuscript writing [11, 12]. Qualitative studies have begun exploring the use of GPT and ChatGPT for conducting various aspects and types of qualitative analysis, from transcription cleaning to theme generation via thematic analysis [13-19]. Table 1 summarizes these studies by describing how GPT and ChatGPT were utilized in the analysis process, the main findings, and the challenges faced during the process. While all these studies are in preprint form and some are awaiting formal peer review, they provide an early glimpse into the feasibility of harnessing ChatGPT as an assistive tool when conducting qualitative analysis.

Whereas these previous papers have focused on the broader use of (Chat)GPT in thematic analysis, its integration into medical research has yet to be investigated. In this viewpoint, we therefore explore the utilization of ChatGPT for thematic analysis in the medical domain, while addressing the unique challenges that arise within a medical context. We begin by assessing the use of ChatGPT for generating codes based on an interview transcript, followed by extracting themes from a list of generated codes. Subsequently, we utilize ChatGPT for cleaning quotes for manuscript preparation. Finally, we use ChatGPT to generate interview transcripts, which may be used for various academic and educational purposes. For each application, we identify the areas where human intervention may still be required.

**Table 1.** Summary of studies that utilized ChatGPT/GPT for different phase(s) of thematic analysis

| Reference | Country | GPT model | Usage of ChatGPT/GPT | Findings | Challenges with ChatGPT/GPT* |
|---|---|---|---|---|---|
| De Paoli (2023, *preprint*) [13] | UK | GPT-3.5 Turbo | Study utilized GPT to conduct thematic analysis, including generating initial codes, searching for themes, reviewing themes, and defining and naming themes. | GPT-3.5 Turbo was able to provide themes with synthetic descriptions. However, some inferred themes that were not considered relevant by human researchers and ChatGPT missed out on themes that were reported by human researchers. | • Interviews had to be divided into chunks due to token limit.<br>• Output is prompt-dependent (e.g., asking for different number of themes produced a different set of themes).<br>• Hallucination (e.g., assigned incorrect code to theme). |
| De Paoli (2023, *preprint*) [14] | UK | GPT-3.5 Turbo | Study utilized GPT to build user personas (i.e., fictional, yet realistic, description of a typical/target user of a product [20]) based on interview transcripts using thematic analysis. | GPT-3.5 Turbo was able to generate two relevant personas based on challenges and needs identified during thematic analysis. | • Biased towards creating specific types of user personas (e.g., mostly middle-aged, from Italy)<br>• Required human intervention to refine codes and themes generated (e.g., generated code with truncated quote) |
| Gao et al. (2023, *preprint*) [15] | Singapore and USA | GPT-3.5 | Study explored the functionality of CollabCoder (a data management prototype incorporating GPT-3.5) in assisting with open coding, iterative discussions and development of codebook. | Participants valued GPT-3.5 for reducing cognitive burden during coding, but some participants cited that summaries generated by GPT-3.5 are too detailed and not relevant. | • Does not consider research questions or intended direction of analysis if not explicitly instructed. |
| Mesec (2023, *preprint*) [16] | Slovenia | Information not available | Study utilized ChatGPT to conduct qualitative analysis using grounded theory method. | ChatGPT was able to summarize and balance opposing ideas but tended to express ideas using descriptive terms at a lower level of abstraction compared to human researcher. | • Hallucination (e.g., made up information in summary of texts)<br>• Codes inadequately captured content of transcript<br>• Unproductive repetition of output<br>• Inappropriate use of terms (e.g., “we can form some qualitative analyses”) |
| Tabone & de Winter (2023, *preprint*) [17] | Netherlands | GPT-3.5 Turbo and GPT-4-0613 | Study utilized GPT to i) conduct sentiment analysis, ii) provide meta-summaries of interviews and iii) identify differences between two think-aloud transcripts. | Ratings (r=.98) and summaries generated with GPT-3.5 were strongly correlated or generally in-line with those generated by human researchers. GPT-3.5 was also able to summarize descriptive differences between two transcripts. | • Prompt dependent (e.g., modified prompt increased correlation of ratings)<br>• Summary by GPT-4.0 was richer and touches on more facets than GPT-3.5 Turbo, however, some topics that were identified did not emerge in content analysis conducted by humans. |
| Taylor (2023, *preprint*) [18] | USA | Information not available | Study utilized ChatGPT to clean interview transcripts after using AI-assisted method to transcribe interviews. | ChatGPT cleaned redundant words and sentence fragments, but transcripts were more difficult to read due to ChatGPT connecting sentence fragments, which resulted in longer words per sentence. | • Word and syntax errors remained in several transcriptions.<br>• Quality of transcription cleaning dependent on speech of speaker (e.g., clarity, pauses, filler words)<br>• Limited input word count (500-600 words as of March 2023) |
| Xiao et al. (2023, *preprint*) [19] | Canada and France | GPT-3 | Study utilized GPT to conduct deductive coding using an expert-developed codebook. | GPT-3 achieved fair to substantial agreement with human researcher (Cohen's $\kappa$ = 0.38-0.61). | • The model occasionally produced incorrect labels. |

Note: GPT, Generative Pre-trained Transformer; UK, The United Kingdom; USA; The United States of America

## Utilizing ChatGPT in Thematic Analysis

Given ChatGPT's ability to handle large textual data and provide sets of meaningful codes and themes, as demonstrated by De Paoli [13], ChatGPT has the potential to improve the efficiency of the thematic analysis process. As thematic analysis is typically conducted using a cyclical and iterative manner (e.g., data collection and data analysis should occur concurrently, with insights from the analysis informing subsequent rounds of data collection and vice versa) [21], being able to digest and process large amounts of information efficiently (e.g., through requesting a summary of an interview or generating an initial set of codes to help breakdown a transcript) can be helpful to researchers streamline this cyclical process. Beyond that, efficiency during this process may translate into lower research costs due to fewer hours spent on analysis.

To illustrate the various ways ChatGPT can be employed for thematic analysis, we used ChatGPT-3.5, which is free of charge. We have also used ChatGPT-4.0, which currently requires a fee, to see whether the results would improve when using a newer version of ChatGPT. Analyses were done on a transcript of the first episode of "Diabetes Discussion – A Diabetes UK Podcast", in which two guests share their experiences of living with diabetes [22].

### Coding the transcript

Our starting point is to investigate the capability of ChatGPT to code transcripts directly. To this end, we made the following request in ChatGPT:

The following is a transcript of an interview for a scientific paper focused on experiences of living with diabetes. Label the text by codes as is done in thematic analysis. Give the codes in the following format:

CODE
- First words of the sentence(s) that was/were labeled with this code

[transcript]

A section of the coded transcript by ChatGPT-3.5 is displayed in Figure 2. ChatGPT-3.5 successfully identified multiple codes in the transcript within a single answer, and the corresponding codes match the textual content. However, as the transcript progresses (Supplementary Figure 1), the coding by ChatGPT-3.5 becomes less detailed. While the subsequent answer contains multiple topics, ChatGPT-3.5 only assigned a single code, resulting in a loss of information in the coded output. Additionally, the second interview question (see Supplementary Figure 1) is also coded, which is not a common practice in thematic analysis.

Using ChatGPT-4.0 for the same analysis results in a noticeable improvement in the output (Supplementary Figure 2). Not only does ChatGPT-4.0 ignore the transcript of the sections corresponding to the interviewer, but it also captures more details of the transcript. ChatGPT-4.0 assigned five different codes for the second question (as opposed to one by ChatGPT-3.5), resulting in a set of codes that give a more complete picture of the transcript.

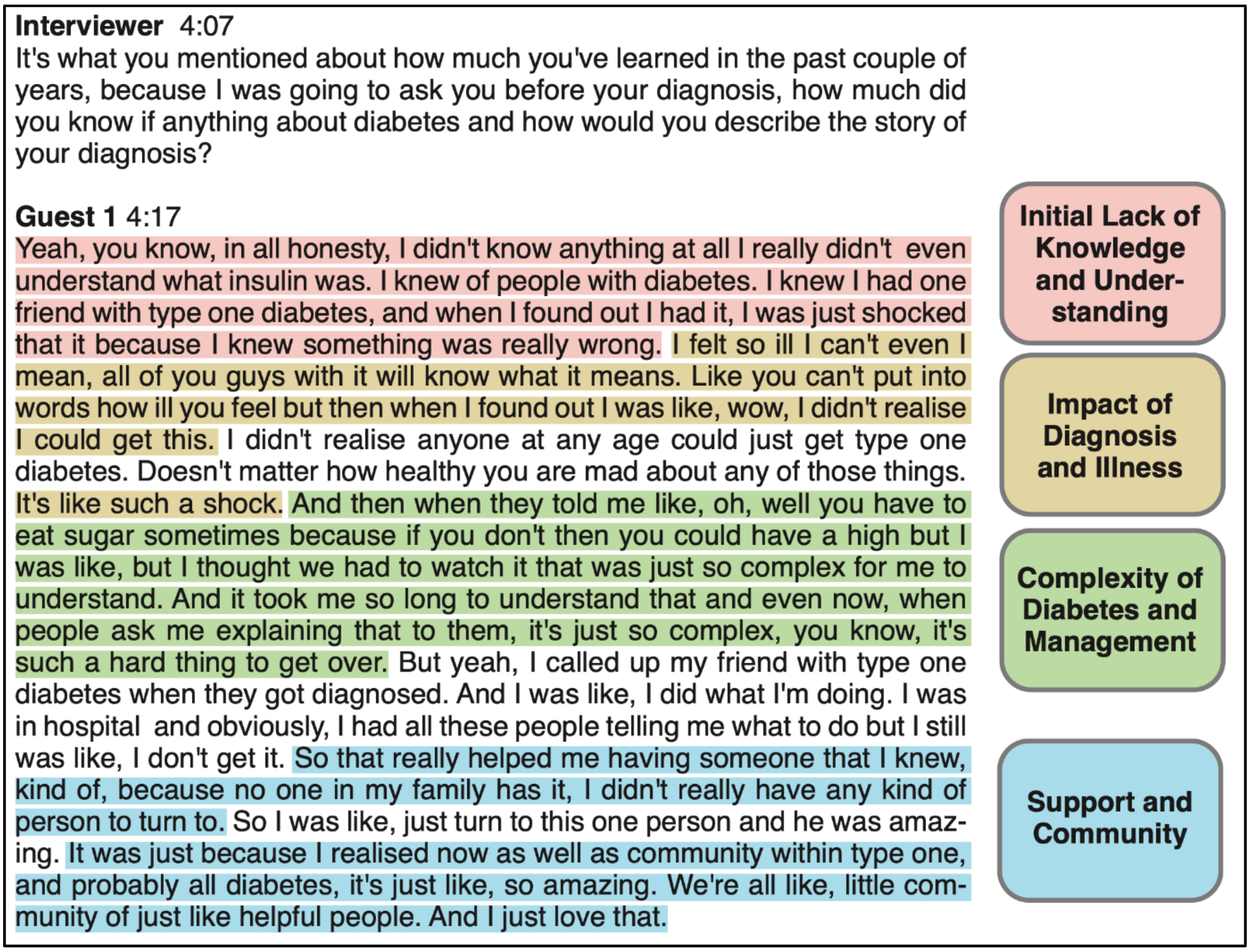

**Figure 2:** Transcript coded by ChatGPT-3.5. A larger portion of the coded transcript is given in Supplementary Figure 1.

ChatGPT demonstrates a promising ability to code transcripts, but its performance depends on the GPT model utilized. Consequently, a human researcher is still necessary to review the codes generated and ensure that the codes appropriately capture all essential data. Furthermore, while the codes generated by ChatGPT sufficiently describe important concepts within the transcript, initial codes generated by human researchers during data analysis may continue to evolve to become more specific or reworded to better capture the data as new transcripts are being analyzed. Given ChatGPT's limited context window, ChatGPT may not remember the full text of an interview transcript and/or the codes it previously generated, resulting in a loss of contextual understanding. For these reasons, a human researcher will be required to consolidate the codes generated to ensure that the codes adequately capture concepts or patterns of interest within the context of the whole dataset.

**Extracting themes from codes**

Another way that ChatGPT may be utilized is to extract the themes and subthemes from the generated codes. These codes may be either obtained from ChatGPT or a human analyzer. We used the following request:

The following codes were obtained via coding of a transcript. Please identify the overarching themes and subthemes as is done in thematic analysis. These themes should have as little overlap as possible, and will be used in a scientific paper focused on experiences of living with diabetes.  Use the following format:

**THEME* *
*Subtheme*:
- codes that belong to this subtheme
*Subtheme*:
- codes

**THEME**
*Subtheme*
...

[list of codes]

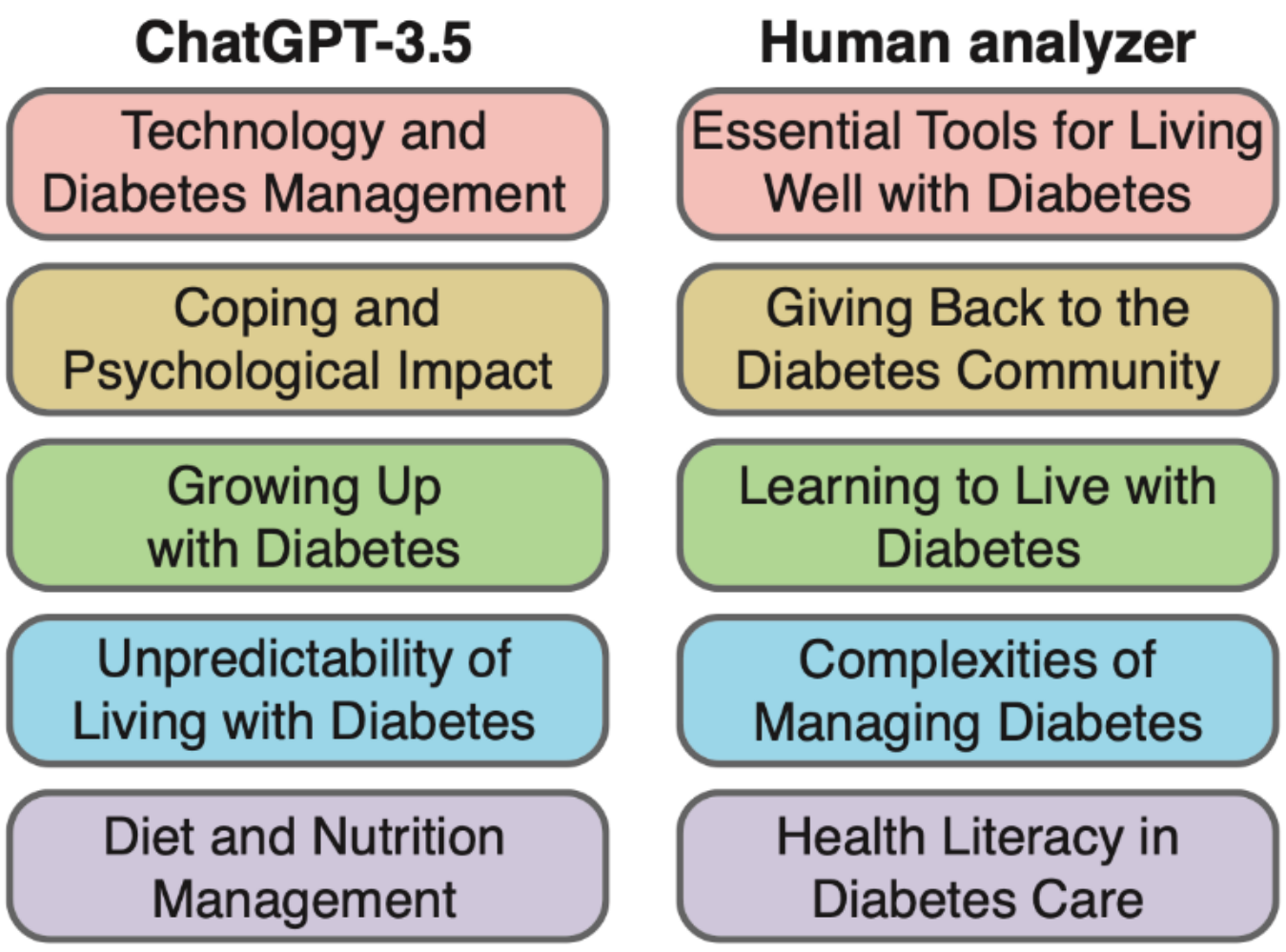

**Figure 3:** Themes identified by ChatGPT-3.5 (left) and the human analyzer (right) from 81 codes.

The resulting themes are shown in Figure 3; the subthemes are tabulated in Supplementary Table 1. The themes and subthemes were derived from 81 codes obtained through coding of the Spotify transcript. Both ChatGPT and the human analyzer identified five unique themes, though their analyses had notable differences. For example, ChatGPT identified the “Diet and Nutrition Management" theme, a subject that the human analyst neither classified as a theme nor a subtheme (Supplementary Table 1).

We conducted several rounds of analysis with ChatGPT-3.5 by resubmitting the same prompt, which led to interesting variations in the identified themes. For example, in the second analysis round, ChatGPT identified the "Pregnancy and Diabetes" theme (Supplementary Table 1), whereas pregnancy only emerged as a subtheme in ChatGPT's first analysis round. The human analyzer, in contrast, did not identify pregnancy as a theme nor as a subtheme in their analysis.

It is impossible to state which of these themes more accurately reflects the essence of the interview, as there is no absolute truth in thematic analysis. Instead, we see the identification of different themes as an advantage. After all, a greater diversity in themes indicates that the codes were interpreted from different angles, thereby adding more layers to the overall analysis. In this framework, ChatGPT should be viewed as an additional team member when doing thematic analysis by offering fresh perspectives and proposing alternative interpretations of the identified codes.

Similar to the coding process, all themes and subthemes generated by ChatGPT should still be reviewed by human researchers to ensure that the themes and subthemes generated are aligned with the research question(s) and essential data have been appropriately captured by the themes.

**Cleaning quotes**

Another potential use of ChatGPT is to clean quotes from the interview transcript for manuscript preparation. We used the following request:

> Clean the following transcript so it may be used as a direct quote for an academic paper:
>
> [quote]
>
> Omit all text that is not essential for the main message. Any altered or inserted words must be shown between square brackets, "[ ]", and omitted text must be replaced by three dots, "…".

**Original transcript**

"Yeah, I mean, I, I think probably we all do, I have a dietician, who does call me quite regularly and discusses the foods I eat. And obviously, they've got access to my CGM. And if I have a spike in any point of the day, they're like, What did you eat here? And I'm like, I don't know. But um, I can guess quite often. And I'm like, oh, yeah, that was porridge again, you know, something like that. So I guess I've just learned myself, what triggers me to go too high, or what doesn't need as much insulin kind of thing."

**ChatGPT-3.5**

"I have a dietician who regularly reviews my diet and monitors my CGM data. If I experience a glucose spike, they inquire about the corresponding meal. While I may not always recall precisely, I can often make an educated guess. Over time, I've personally learned my own triggers for hyperglycemia and instances where reduced insulin is required."

**Original transcript**

"Yeah, I mean, I, I think probably we all do, I have a dietician, who does call me quite regularly and discusses the foods I eat. And obviously, they've got access to my CGM. And if I have a spike in any point of the day, they're like, What did you eat here? And I'm like, I don't know. But um, I can guess quite often. And I'm like, oh, yeah, that was porridge again, you know, something like that. So I guess I've just learned myself, what triggers me to go too high, or what doesn't need as much insulin kind of thing."

**ChatGPT-4.0**

"I have a dietician who … discusses the foods I eat. They have access to my CGM. If I have a spike … they ask, 'What did you eat here?' … I've learned … what triggers me to go too high, or what doesn't need as much insulin."

**Figure 4.** Direct quotes generated by ChatGPT-3.5 (left) and ChatGPT-4.0 (right). The yellow highlights indicate the text that has been included in the quote, whereas the red highlights indicate the text that has been rephrased by ChatGPT without proper use of square brackets.

As shown in Figure 4, ChatGPT-3.5 struggled to clean the quote for manuscript writing purposes. A substantial portion of the text underwent revision without proper use of square brackets, despite our explicit request to denote any modifications or insertions. Additionally, sections of the text were interpreted by ChatGPT-3.5, which deviated from our initial instructions. In contrast, ChatGPT-4.0 performed the task more effectively (Figure 4). The generated quote contains only one word that should have been added between square brackets, while the remainder of the quote is indeed correctly extracted from the original transcript. Nevertheless, depending on how the quote is utilized to describe the data, one may argue that too much text has been omitted from the original transcript, filtering out the more personal viewpoints or emotional content from the final quote. Accordingly, quotes cleaned by ChatGPT should be reviewed by human researchers to determine if the quotes provide sufficient information and context to further elaborate the theme or subtheme discussed. As cleaning up quotes during manuscript preparation is less time-consuming than coding and theme generation, researchers should consider if the steps needed to generate a clean quote outweigh the manual process.

### Creating transcripts

Finally, we have explored the potential of ChatGPT to generate transcripts of interviews. To this end, we made the following request to ChatGPT:

Generate a transcript of an interview for a scientific study with a diabetes patient about their experience with living with diabetes. Required length: ~1500 words. Use the following format:

Interviewer mm:ss
[Question 1]

Patient 1 mm:ss
[Answer]

Include stop words, pauses, and words such as "I think..,", "uhh", and "I mean...", so it looks like an authentic transcript.

This request was made without any additional instructions, such as an example transcript or a list of themes or codes. A snapshot of the generated transcript is shown in Textbox 1. The resulting transcript matches the requested format and uses the stopwords that were asked for, making it similar to an actual conversation between two people. However, some differences exist between the generated transcripts and those from real-life interviews. The ChatGPT-generated transcript has a very direct question-answer structure, with all answers being on-point and of similar length. In contrast, a real-life interview is often more organic. For example, the interviewee may not understand the question and give answers of various lengths, and the interviewer may go in more depth before moving to the next question.

**Interviewer: 00:30**
[...] Can you share what your initial thoughts and feelings were when you received the diagnosis?

**Patient 1: 00:36**
Oh, absolutely. Well, I remember feeling, you know, a mix of emotions. I mean, first there was this sense of disbelief, like, "Is this really happening to me?" And, uh, I guess there was some fear too, you know, about what it would mean for my future. I think I was also a bit confused about how it all worked, you know, managing blood sugar levels and all that.

**Interviewer: 00:54**
It's completely understandable to feel that way. Managing diabetes can be complex. How has your daily routine changed since your diagnosis?

**Patient 1: 01:02**
Oh, my daily routine has changed quite a bit, actually. I mean, now I have to, you know, check my blood sugar levels regularly, and, umm, I have to be more mindful of what I eat. So, I've had to make some adjustments to my diet, like cutting back on sugary stuff and, uhh, watching my portion sizes. And, uh, I've also started incorporating, you know, regular exercise into my routine. It's been a bit challenging to juggle everything, but I think I'm getting the hang of it.

**Textbox 1.** Snapshot of the transcript generated by ChatGPT-3.5. The full transcript is given in the Supplementary Information.

Despite these differences, the ChatGPT-generated transcripts have multiple advantages. By slightly adjusting the ChatGPT request, the generated transcripts may focus on any topic of interest or feature different respondent characteristics. This adaptability of ChatGPT allows for the quick generation of a large and diverse collection of transcripts. Furthermore, generating transcripts with ChatGPT eliminates any privacy- and confidentiality concerns, which is a common issue with actual interviews.

Given these advantages, we envision several potential applications for the ChatGPT-generated transcripts. Firstly, these transcripts may be used as instruction material for students learning thematic analysis. A second approach worth investigating is to use the generated transcripts as a training set for NLP models, particularly in topic modeling. As real-life transcripts are often limited or hard to get, ChatGPT offers a practical way to expand the dataset, thereby exposing the model to a larger diversity of text.

In short, while ChatGPT-generated transcripts are not (yet) a perfect substitute for real ones, they offer a promising alternative for various academic and educational applications.

## Challenges when using ChatGPT for thematic analysis

While ChatGPT is a promising assistive tool for thematic analysis, previous studies have identified challenges when working with ChatGPT (see Table 1) [13-19]. Major challenges relevant to thematic analysis include hallucination (i.e., responses produced by the system that is not justified by the data used), the output being prompt-dependent (e.g., prompts requesting the same output but phrased differently will lead to different outputs), and missing themes or codes previously reported by researchers [13]. Similarly, we encountered several challenges when utilizing ChatGPT to conduct thematic analysis. When working with patient data, the primary concern is data confidentiality. Inputs to ChatGPT may be used as training data to improve their services, and network activities may be shared with third parties [23]. For this reason, uploading sensitive information, such as patient interview transcripts, to ChatGPT should be avoided. This precaution restricts the use of ChatGPT for coding (i.e., Phase 2 in Figure 1) unless the transcript holds no confidential information.

A more practical challenge is that ChatGPT has a word limit for each prompt, which may prevent users from inputting full transcripts and very long lists of codes or quotes. One potential solution is to split the input. However, as discussed above, ChatGPT has a limited context window, so it may forget the earlier parts of the interview transcript and/or codes it previously generated. As a result, ChatGPT may not be able to adequately capture patterns of ideas or concepts within the context of the whole dataset. When coding, researchers consider existing knowledge (e.g., the research question or current information about the topic), knowledge obtained through data collection (e.g., interviews and field notes), and existing codes from previously analyzed transcripts. Without further information beyond the input, ChatGPT may adopt a narrower lens and generate results that are highly specific to a singular transcript. Accordingly, at this point in time, it is still essential for human researchers to collate codes generated for each transcript and review them within the context of the study.

In the context of understanding text, ChatGPT, though advanced, may not capture every nuance that a human analyst would pick up [24]. In certain instances, ChatGPT may overlook underlying emotions or implicit themes that would otherwise be evident to human analyzers. It is thus important to review the output of ChatGPT to ensure that all essential aspects of the transcript are captured in its thematic analysis. Beyond that, we also found that ChatGPT sometimes excludes existing codes or introduces new codes when generating themes. Hence, it is advisable to double-check whether all codes have been correctly assigned to the identified themes and subthemes.

Finally, ChatGPT may give different answers to the same questions, leading to non-reproducible results. Yet, in the context of thematic analysis, we do not see this variability as a drawback, as humans would also generate different results when doing thematic analysis. Instead, the different responses by ChatGPT may be seen as an opportunity because it may provide new insights that were not captured during the first round of thematic analysis.

## Conclusions and Recommendations

ChatGPT has the potential to enrich the space of qualitative research. In our investigation, ChatGPT demonstrated its ability to code interview transcripts, generate themes from a list of codes, clean quotes for manuscript preparation, and generate unique transcripts for education and training purposes. Nevertheless, limitations such as the inability to manage multiple transcripts and not fully capturing nuanced data essential to the research question necessitate the involvement of human researchers to collate and review the output generated. At this stage, ChatGPT requires human-AI collaboration, where researchers have to remain in the loop to intervene when necessary [13, 15]. We present Figure 5 to show the opportunities available for ChatGPT to assist in thematic analysis and areas where human involvement is still required.

Given the need for considerable interaction between ChatGPT and human researchers, it will be more valuable to recognize ChatGPT as an additional member of the analysis team, contributing to researcher triangulation by adding to knowledge building and sensemaking rather than a replacement for human researchers. However, with the ongoing progress in the field of natural language processing, the role of ChatGPT in qualitative research will evolve rapidly. This fast-paced development, in combination with the growing use of ChatGPT in research, necessitates further discussions regarding the use of ChatGPT in qualitative research. For example, how should ChatGPT's contribution be acknowledged, and what are the best practices regarding prompt formulation? Another important point of consideration is the confidentiality of the data, especially when working with patient data such as interview transcripts. The recent ChatGPT data breach in March 2023 should encourage researchers to remain mindful of the implications when working with AI tools that store data, regardless of purpose [25].

In summary, ChatGPT has the potential to function as a valuable tool during analysis, enhancing the efficiency of the thematic analysis and offering additional insights into the qualitative data. While the current viewpoint remains an exercise to investigate the potential feasibility of using ChatGPT for thematic analysis, findings from the investigation can serve as a starting point for future studies that intend to further push the boundaries of AI involvement within qualitative research.

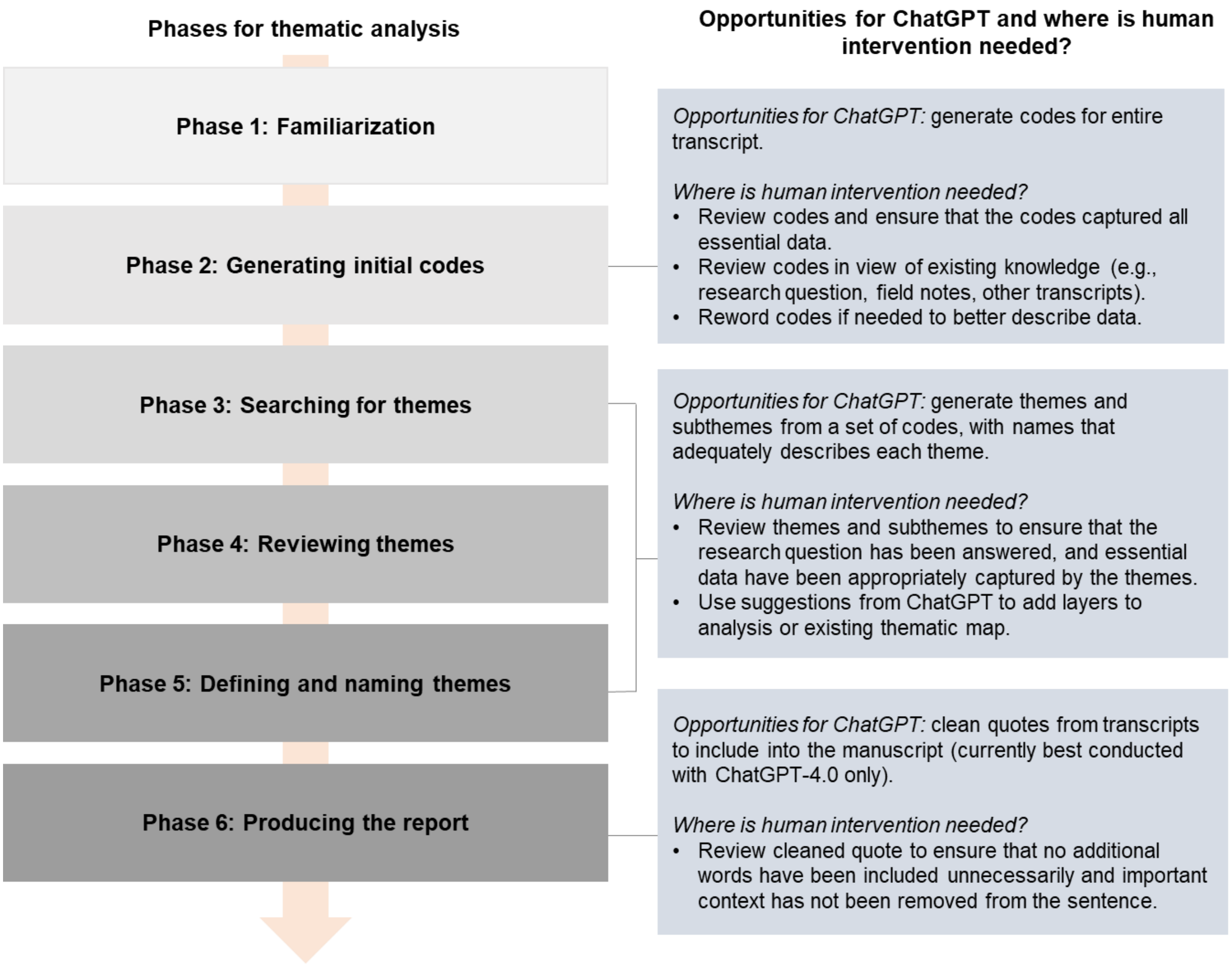

**Figure 5.** Opportunities for the utilization of ChatGPT during the thematic analysis process and where human intervention is needed.

## Acknowledgement

This work was funded by the DFM Research Capabilities Building Budget under the project "Technology and Compassion: Improving Patient Outcomes Through Data Analytics and Patients' Voice in Primary Care" [NUHSRO/2022/049/NUSMed/DFM].

## Authors' Contribution

VVL, SCCL, LHG and JMV conceptualized and outlined the investigation. VVL and SCCL wrote the original draft, and all authors reviewed and edited the manuscript. LHG and JMV supervised the investigation. VVL and SCCL are considered co-first authors.

## Conflicts of Interest

None declared.

## Abbreviations

AI: Artificial Intelligence
GPT: Generative Pre-trained Transformers
NLP: natural language processing
UK: The United Kingdom
USA: The United States of America

# Supplementary Information

**Interviewer** 4:07
It's what you mentioned about how much you've learned in the past couple of years, because I was going to ask you before your diagnosis, how much did you know if anything about diabetes and how would you describe the story of your diagnosis?

**Guest 1** 4:17
Yeah, you know, in all honesty, I didn't know anything at all I really didn't even understand what insulin was. I knew of people with diabetes. I knew I had one friend with type one diabetes, and when I found out I had it, I was just shocked that it because I knew something was really wrong. I felt so ill I can't even I mean, all of you guys with it will know what it means. Like you can't put into words how ill you feel but then when I found out I was like, wow, I didn't realise I could get this. I didn't realise anyone at any age could just get type one diabetes. Doesn't matter how healthy you are mad about any of those things. It's like such a shock. And then when they told me like, oh, well you have to eat sugar sometimes because if you don't then you could have a high but I was like, but I thought we had to watch it that was just so complex for me to understand. And it took me so long to understand that and even now, when people ask me explaining that to them, it's just so complex, you know, it's such a hard thing to get over. But yeah, I called up my friend with type one diabetes when they got diagnosed. And I was like, I did what I'm doing. I was in hospital and obviously, I had all these people telling me what to do but I still was like, I don't get it. So that really helped me having someone that I knew, kind of, because no one in my family has it, I didn't really have any kind of person to turn to. So I was like, just turn to this one person and he was amazing. It was just because I realised now as well as community within type one, and probably all diabetes, it's just like, so amazing. We're all like, little community of just like helpful people. And I just love that.

**Interviewer** 5:45
It's really interesting to hear your stories as someone diagnosed as a young adult. And you're right, I think it's such a steep learning curve for anybody, no matter what age they are. But Katie, much like me, you've grown up with diabetes always being a part of your life. So what was it like for you and your family adjusting to life with diabetes as you were growing up?

**Guest 2** 6:01
Yeah, so I was diagnosed just before I turned two, so I can't remember anything about my diagnosis. And it was very definitely, you know, my parents that had to deal with that initial shock, which I just think parents of young children with diabetes are absolutely amazing. I'm in awe of what they're able to do and how much they're able to take on. I think as you grow up, at each life stage, you kind of learn a little bit more about diabetes, and how to take on a little bit more responsibility. So I think when I was around seven, I started doing my own injections, because I had to be doing that for school. So that was a big change, that was a lot of responsibility for taking on as a child. And then as you get older, as you get through your teenage years, that becomes more and more important to have that responsibility and independence. But I've always been really supported in that I have always had friends with diabetes, I've been really fortunate in the peer support. As you're saying the community, it took me a long time to find the online community. But as a child, I was part of lots of different play groups of young children with diabetes and friends within clinics, I'd see every three months and we'd have a great time catching up and playing with all of those toys in the clinic waiting area. So that was always really helpful. And then my family who have just been amazing, helping me grow up with this and helping me take on that independence. As I've grown up with diabetes as well.

Initial Lack of Knowledge and Understanding

Impact of Diagnosis and Illness

Complexity of Diabetes and Management

Support and Community

Diagnosis and Learning Curve (Young Adult)

Growing Up with Diabetes (Diagnosed in Childhood)

**Supplementary Figure 1 |** Transcript coded by ChatGPT-3.5

**Interviewer** 4:07
It's what you mentioned about how much you've learned in the past couple of years, because I was going to ask you before your diagnosis, how much did you know if anything about diabetes and how would you describe the story of your diagnosis?

**Guest 1** 4:17
Yeah, you know, in all honesty, I didn't know anything at all I really didn't even understand what insulin was. I knew of people with diabetes. I knew I had one friend with type one diabetes, and when I found out I had it, I was just shocked that it because I knew something was really wrong. I felt so ill I can't even I mean, all of you guys with it will know what it means. Like you can't put into words how ill you feel but then when I found out I was like, wow, I didn't realise I could get this. I didn't realise anyone at any age could just get type one diabetes. Doesn't matter how healthy you are mad about any of those things. It's like such a shock. And then when they told me like, oh, well you have to eat sugar sometimes because if you don't then you could have a high but I was like, but I thought we had to watch it that was just so complex for me to understand. And it took me so long to understand that and even now, when people ask me explaining that to them, it's just so complex, you know, it's such a hard thing to get over. But yeah, I called up my friend with type one diabetes when they got diagnosed. And I was like, I did what I'm doing. I was in hospital and obviously, I had all these people telling me what to do but I still was like, I don't get it. So that really helped me having someone that I knew, kind of, because no one in my family has it, I didn't really have any kind of person to turn to. So I was like, just turn to this one person and he was amazing. It was just because I realised now as well as community within type one, and probably all diabetes, it's just like, so amazing. We're all like, little community of just like helpful people. And I just love that.

**Interviewer** 5:45
It's really interesting to hear your stories as someone diagnosed as a young adult. And you're right, I think it's such a steep learning curve for anybody, no matter what age they are. But Katie, much like me, you've grown up with diabetes always being a part of your life. So what was it like for you and your family adjusting to life with diabetes as you were growing up?

**Guest 2** 6:01
Yeah, so I was diagnosed just before I turned two, so I can't remember anything about my diagnosis. And it was very definitely, you know, my parents that had to deal with that initial shock, which I just think parents of young children with diabetes are absolutely amazing. I'm in awe of what they're able to do and how much they're able to take on. I think as you grow up, at each life stage, you kind of learn a little bit more about diabetes, and how to take on a little bit more responsibility. So I think when I was around seven, I started doing my own injections, because I had to be doing that for school. So that was a big change, that was a lot of responsibility for taking on as a child. And then as you get older, as you get through your teenage years, that becomes more and more important to have that responsibility and independence. But I've always been really supported in that I have always had friends with diabetes, I've been really fortunate in the peer support. As you're saying the community, it took me a long time to find the online community. But as a child, I was part of lots of different play groups of young children with diabetes and friends within clinics, I'd see every three months and we'd have a great time catching up and playing with all of those toys in the clinic waiting area. So that was always really helpful. And then my family who have just been amazing, helping me grow up with this and helping me take on that independence. As I've grown up with diabetes as well.

**Previous Knowledge About Diabetes**

**Initial Shock & Reaction**

**Complexity & Misunderstanding**

**Seeking Personal Experience for Guidance**

**Value of Diabetes Community**

**Early Age Diagnosis**

**Parental Challenge & Admiration**

**Gradual Learning & Increased Responsibility**

**Peer Support & Community Importance**

**Family Support & Growth with Diabetes**

**Supplementary Figure 2 |** Transcript coded by ChatGPT-4.0

**Supplementary Information 1: Transcript generated by ChatGPT-3.5**

**Interviewer: 00:00**
Good morning. Thank you for participating in this interview for our scientific study on the experience of living with diabetes. Let's start with a basic question: Can you tell me when you were diagnosed with diabetes and what type of diabetes you have?

**Patient 1: 00:12**
Sure, uhh, I was diagnosed about five years ago. It was, umm, I think it was in my early 40s. I have type 2 diabetes. Yeah, I mean, it was a bit of a shock, you know? I didn't really see it coming.

**Interviewer: 00:30**
I can imagine that was quite surprising. Can you share what your initial thoughts and feelings were when you received the diagnosis?

**Patient 1: 00:36**
Oh, absolutely. Well, I remember feeling, you know, a mix of emotions. I mean, first there was this sense of disbelief, like, "Is this really happening to me?" And, uh, I guess there was some fear too, you know, about what it would mean for my future. I think I was also a bit confused about how it all worked, you know, managing blood sugar levels and all that.

**Interviewer: 00:54**
It's completely understandable to feel that way. Managing diabetes can be complex. How has your daily routine changed since your diagnosis?

**Patient 1: 01:02**
Oh, my daily routine has changed quite a bit, actually. I mean, now I have to, you know, check my blood sugar levels regularly, and, umm, I have to be more mindful of what I eat. So, I've had to make some adjustments to my diet, like cutting back on sugary stuff and, uhh, watching my portion sizes. And, uh, I've also started incorporating, you know, regular exercise into my routine. It's been a bit challenging to juggle everything, but I think I'm getting the hang of it.

**Interviewer: 01:25**
It sounds like you've made some positive changes to manage your diabetes. Can you tell me about any specific challenges you've faced in your daily life as a result of living with diabetes?

**Patient 1: 01:34**
Oh, definitely. I mean, there have been some challenges for sure. Umm, one thing that comes to mind is, you know, eating out or social gatherings. It can be a bit awkward sometimes, trying to make healthy choices while also, you know, not making a big deal out of it. And, uh, I've had my moments of frustration too, especially when my blood sugar levels, like, they just don't seem to cooperate, no matter what I do. It's a bit frustrating, I mean, you know, I'm trying my best here.

**Interviewer: 01:56**
I can see how those situations can be tough. How about the emotional aspect of living with diabetes? Have you experienced any emotional ups and downs related to your condition?

**Patient 1: 02:06**
Oh, definitely. I mean, there have been times when I've felt a bit down about it. You know, I think it's natural to have those moments of, like, "Why me?" or feeling, you know, a bit overwhelmed. And then there are moments when I'm proud of myself, you know, when I've managed to keep my blood sugar levels in check or I've hit my exercise goals. It's a bit of a rollercoaster, I mean, emotionally.

**Interviewer: 02:25**
Absolutely, it's important to acknowledge both the challenges and the victories. How has your support system been throughout this journey? Have you received support from family, friends, or healthcare professionals?

**Patient 1: 02:35**
Yeah, I'm really grateful for the support I've received. Umm, my family has been, you know, quite understanding and they've tried to make, like, healthier meals when I'm around. And my friends, well, they've been pretty great too, you know, making sure there are options that I can eat if we go out. And my doctor, uh, they've been really helpful in explaining things to me and, you know, answering my questions. So, yeah, I think having a good support system makes a big difference.

**Interviewer: 02:57**
It sounds like you have a strong network to lean on. Looking back, what advice would you give to someone who has recently been diagnosed with diabetes?

**Patient 1: 03:05**
I would say, you know, take it one step at a time. It can be overwhelming at first, but you'll learn as you go. And, uh, don't be afraid to ask questions, whether it's from your doctor or, you know, support groups online. And, umm, also, I think it's important to be kind to yourself. There will be days when things don't go as planned, and that's okay. You're doing your best, and that's what matters.

**Interviewer: 03:23**
Great advice. Finally, can you share any positive aspects that have come from your experience with diabetes? Have there been any unexpected silver linings?

**Patient 1: 03:32**
Oh, yeah, definitely. I mean, I think, you know, having diabetes has made me more conscious of my overall health. I've, umm, I've learned a lot about nutrition and exercise that I might not have paid as much attention to before. And, you know, it's also brought me closer to some people in my life. Like, my friends and family, they've really shown their support and it's been, you know, heartwarming. So, yeah, I think there have been some positive aspects despite the challenges.

**Interviewer: 03:54**

That's wonderful to hear. Thank you so much for sharing your experiences with us today. Your insights will certainly contribute to our understanding of living with diabetes.

**Patient 1: 04:02**
You're welcome. I'm glad I could help. If my experiences can, you know, help someone else going through the same thing, then that's great. Uhh, thanks for having me.

**Interviewer: 04:10**
Thank you again, and take care.

**Supplementary Table 1.** Themes and subthemes identified by ChatGPT-3.5 during the first analysis round (left) and second analysis round (middle), and the human analyzer (right) from 81 codes.

| ChatGPT-3.5: Analysis 1 | | ChatGPT-3.5: Analysis 2 | | Human analyzer | |
|---|---|---|---|---|---|
| **Themes** | **Subthemes** | **Themes** | **Subthemes** | **Themes** | **Subthemes** |
| Technology and Diabetes Management | - Use of Diabetes Technology<br>- Healthcare Support and Education | Diabetes Management Strategies | - Technological Aids<br>- Diabetes Education and Peer Support<br>- Dietary Management<br>- Emotional and Mental Aspects | Essential Tools for Living Well with Diabetes | - Preparation and Planning<br>- Supportive Network and healthcare Team<br>- Continuous Self-learning<br>- Access to Digital Health Tool |
| Coping and Psychological Impact | - Psychological and Emotional Struggles<br>- Stigma and Normalization | Stigma and Normalization | - Battling Stigma<br>- Advocating for Normalcy | Giving Back to the Diabetes Community | - Tackling Stigma and Normalizing Diabetes |
| Growing Up with Diabetes | - Diagnosis at Young Age<br>- Independence and Responsibility<br>- Impact on Social Life and Education | Impact of Diabetes Diagnosis on Life | - Shock and Lack of Knowledge<br>- Supportive Environment and Education<br>- Becoming Independent with Age<br>- Impact on Relationships | Learning to Live with Diabetes | - Parent's Role in child's diagnosis<br>- Able to successfully adapt to living with diabetes |
| Unpredictability of Living with Diabetes | - Unpredictability of Diabetes Management<br>- Challenges in Pregnancy with Diabetes<br>- Diagnosis and Initial Reactions<br>- Support Networks and Community | Unpredictability of Living with Diabetes | - Unpredictability of Diabetes Management<br>- Navigating Uncertainties | Complexities of Managing Diabetes | - Diabetes is Unpredictable<br>- Diabetes Adds a Layer of Challenge to Daily Life<br>- Struggle to Find Suitable Healthcare Providers |
| Diet and Nutrition Management | - Balanced Diet and Food Strategies<br>- Role of Diabetes Education | Pregnancy and Diabetes | - Managing Pregnancy with Diabetes<br>- Challenges of Pregnancy with Diabetes | Health Literacy in Diabetes Care | - Lack of Knowledge about Condition and Management<br>- Struggle to Access Suitable Resources |